\documentclass[conference]{IEEEtran}
\IEEEoverridecommandlockouts

\usepackage{cite}
\usepackage{amsmath,amssymb,amsfonts}
\usepackage{booktabs}
\usepackage{multirow}
\usepackage{array}
\usepackage{graphicx}
\usepackage{siunitx}
\usepackage[table]{xcolor}
\usepackage[most]{tcolorbox}
\usepackage{url}
\usepackage[hidelinks]{hyperref}

\begin{document}

\title{Breaking the 1.58-bit Barrier for Ternary LLMs}

\author{\IEEEauthorblockN{Evangelos Georganas, Alexander Heinecke, Pradeep Dubey}
\IEEEauthorblockA{\textit{Intel Corporation}\\
}}

\maketitle

\begin{abstract}
Ternary Large Language Models (LLM) store every weight as one of three symbols
$\{-1,0,+1\}$, so the cost of a ternary model is conventionally referenced
to the information-theoretic $\log_2 3 \approx 1.585$ bits per weight. The prevailing deployment format packs five ternary weights into one byte (\emph{five-trit packing}), and due to the power-of-two group sizes used in practice this rounds up to $1.625$ bits per weight. This effective storage bit-width treats the
three symbols $\{-1,0,+1\}$ as equiprobable. We measure the actual symbol
distribution of 29 ternary LLM models and find that zeros
account for up to $51.5\%$ of all weights. Motivated by this finding, we introduce BITCOS, a simple distribution-adaptive layout comprised of a dense presence bitmap plus a compacted sign vector, and costs $2 - z$ bits per weight element given a zero density $z$ in the model's weights.
BITCOS stores weights more compactly than the five-trit packing in 26 of the 29 tested models, and reaches $1.485$ bits per weight on the sparsest of them. BITCOS is amenable to efficient unpacking on modern processors and GPUs, and we present optimized unpacking sequences for AVX-512, AVX2 and Intel Xe2 GPUs. Measured against production state-of-the-art ternary matrix-vector multiplication kernels, at the zero densities real-world ternary models exhibit, the realized gain with our proposed layout is up to $1.28\times$. Finally, we illustrate end-to-end LLM inference results on 5 different platforms (client and server CPUs, integrated and discrete Xe2 GPUs) where decode throughput improves by up to $1.18\times$ on CPUs and $1.27\times$ on GPUs.
\end{abstract}

\section{Introduction}
Ternary weight quantization restricts every weight to $\{-1,0,+1\}$ scaled by a group-wise factor~\cite{bitnet158,bitnet2b4t}. Three equiprobable symbols carry $\log_2 3 \approx 1.585$ bits of information, and this theoretical bound is the reference for ternary storage. Nevertheless, in practice the actual storage cost is determined by how those symbols are packed. Five trits (ternary digits) fit in a byte ($3^5 = 243 \le 256$), which approaches the bound at $8/5 = 1.6$ bits per weight. However, deployed implementations quantize in blocks of power-of-two weights like 128, and 128 is not a multiple of 5: a block needs $\lceil 128/5 \rceil = 26$ payload bytes, so the rate stored in practice is $26\times8/128 = 1.625$ bits per weight. Still, this effective storage bit-width treats the
three symbols $\{-1,0,+1\}$ as equiprobable.

\begin{figure}[!ht]
\centering
\includegraphics[width=\columnwidth]{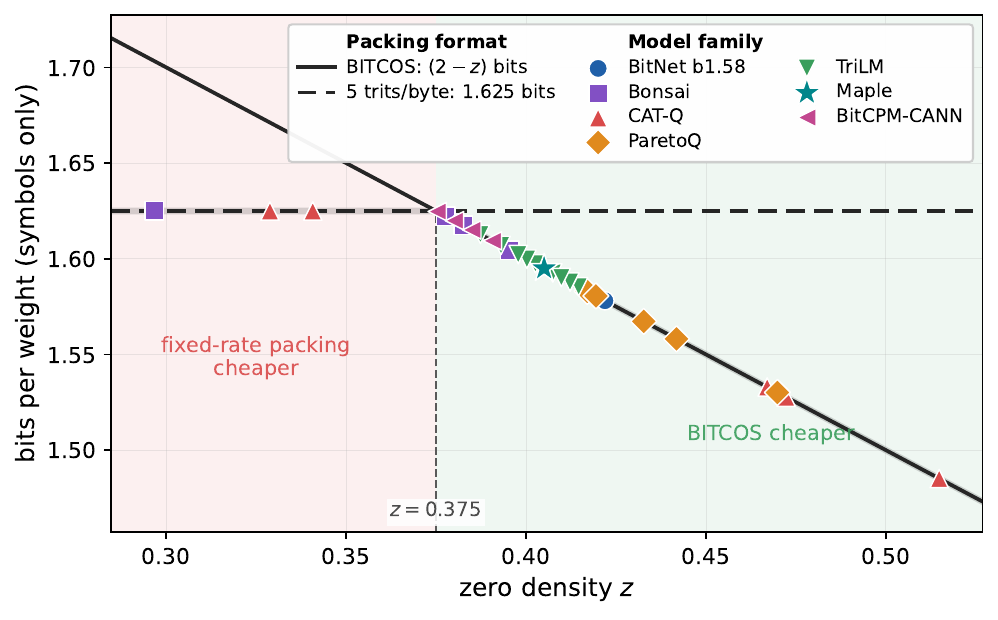}
\caption{Effective bit-widths for various families of ternary LLM models, comparing the five-trit packing and the proposed BITCOS layout. The sloped line represents the BITCOS bit-width $2-z$, while the plateau indicates the fixed five-trit bit-width of $1.625$ bits per weight. Each marker corresponds to a model from Table~\ref{tab:bits}, placed at its measured zero density. Models with zero density $z$ above $0.375$ benefit from the BITCOS layout (green plot area), as such 26 out of the 29 models achieve lower effective bit-width with BITCOS.}
\label{fig:models}
\end{figure}
\begin{table}[!t]
\centering
\scriptsize
\setlength{\tabcolsep}{1pt}
\renewcommand{\arraystretch}{0.9}
\caption{Effective
bit-widths for SOTA ternary LLM models. ``\% 0'' is the measured
zero density $z$. ``Symbols'' column counts the ternary codes
alone and ``$+$scale'' adds the measured 16-bit scale overhead. The symbols-only rate is $2-z$ for BITCOS and for every
model $2.000$ and $1.625$ bits per weight for 2-bit and 5-trit packing respectively.
``red.'' is the size reduction of BITCOS over the corresponding format with scales included. A value $> 1$ means BITCOS stores the model more compactly.}
\label{tab:bits}
\definecolor{famBitNet}{HTML}{1F5FA8}
\definecolor{famBonsai}{HTML}{8250C4}
\definecolor{famCatQ}{HTML}{D94A4A}
\definecolor{famParetoQ}{HTML}{E08A1E}
\definecolor{famTriLM}{HTML}{3A9E5C}
\definecolor{famMaple}{HTML}{00868B}
\definecolor{famBitCPM}{HTML}{C2478F}
\begin{tabular}{cl|c|cc|cc|cc}
\toprule
& & &
\multicolumn{2}{c|}{\textbf{BITCOS}} &
\multicolumn{2}{c|}{\textbf{2-bit packing}} &
\multicolumn{2}{c}{\textbf{5-trit per byte}} \\
 & \textbf{Ternary model} & \textbf{\% 0} & \textbf{Symbols} & \textbf{$+$scale} & \textbf{$+$scale} & \textbf{red.} & \textbf{$+$scale} & \textbf{red.} \\
\midrule
{\color{famBitNet}$\bullet$} & BitNet b1.58 2B4T & 42.19 & 1.578 & ---   & 2.000 & $1.27\times$ & 1.625 & $1.03\times$ \\
\midrule
\multirow{4}{*}{{\color{famBonsai}$\blacksquare$}}
  & Bonsai 1.7B              & 39.89 & 1.601 & 1.726 & 2.125 & $1.23\times$ & 1.750 & $1.01\times$ \\
  & Bonsai 4B                & 37.71 & 1.623 & 1.748 & 2.125 & $1.22\times$ & 1.750 & $1.00\times$ \\
  & Bonsai 8B                & 38.25 & 1.618 & 1.743 & 2.125 & $1.22\times$ & 1.750 & $1.00\times$ \\
  & Bonsai 27B               & 29.66 & 1.703 & 1.828 & 2.125 & $1.16\times$ & 1.750 & $0.96\times$ \\
\midrule
\multirow{5}{*}{{\color{famCatQ}$\blacktriangle$}}
  & CAT-Q Qwen3-1.7B         & 51.48 & 1.485 & 1.610 & 2.125 & $1.32\times$ & 1.750 & $1.09\times$ \\
  & CAT-Q Qwen3-8B           & 46.70 & 1.533 & 1.658 & 2.125 & $1.28\times$ & 1.750 & $1.06\times$ \\
  & CAT-Q Qwen3-30B-A3B      & 32.88 & 1.671 & 1.796 & 2.125 & $1.18\times$ & 1.750 & $0.97\times$ \\
  & CAT-Q Qwen3-32B          & 47.11 & 1.529 & 1.654 & 2.125 & $1.28\times$ & 1.750 & $1.06\times$ \\
  & CAT-Q Qwen3-235B-A22B    & 34.07 & 1.659 & 1.784 & 2.125 & $1.19\times$ & 1.750 & $0.98\times$ \\
\midrule
\multirow{5}{*}{{\color{famParetoQ}$\blacklozenge$}}
  & ParetoQ 125M             & 41.07 & 1.589 & 1.613 & 2.023 & $1.25\times$ & 1.648 & $1.02\times$ \\
  & ParetoQ 350M             & 41.58 & 1.584 & 1.598 & 2.014 & $1.26\times$ & 1.639 & $1.03\times$ \\
  & ParetoQ 600M             & 43.27 & 1.567 & 1.579 & 2.012 & $1.27\times$ & 1.637 & $1.04\times$ \\
  & ParetoQ 1B               & 44.18 & 1.558 & 1.569 & 2.010 & $1.28\times$ & 1.635 & $1.04\times$ \\
  & ParetoQ 1.5B             & 47.10 & 1.529 & 1.537 & 2.008 & $1.31\times$ & 1.633 & $1.06\times$ \\
\midrule
\multirow{9}{*}{{\color{famTriLM}$\blacktriangledown$}}
  & TriLM 99M                & 40.97 & 1.590 & 1.618 & 2.027 & $1.25\times$ & 1.652 & $1.02\times$ \\
  & TriLM 190M               & 40.67 & 1.593 & 1.611 & 2.018 & $1.25\times$ & 1.643 & $1.02\times$ \\
  & TriLM 390M               & 40.79 & 1.592 & 1.606 & 2.014 & $1.25\times$ & 1.639 & $1.02\times$ \\
  & TriLM 560M               & 40.73 & 1.593 & 1.604 & 2.011 & $1.25\times$ & 1.636 & $1.02\times$ \\
  & TriLM 830M               & 40.54 & 1.595 & 1.604 & 2.009 & $1.25\times$ & 1.634 & $1.02\times$ \\
  & TriLM 1.1B               & 40.39 & 1.596 & 1.605 & 2.009 & $1.25\times$ & 1.634 & $1.02\times$ \\
  & TriLM 1.5B               & 40.21 & 1.598 & 1.606 & 2.008 & $1.25\times$ & 1.633 & $1.02\times$ \\
  & TriLM 2.4B               & 39.70 & 1.603 & 1.610 & 2.007 & $1.25\times$ & 1.632 & $1.01\times$ \\
  & TriLM 3.9B               & 38.70 & 1.613 & 1.618 & 2.005 & $1.24\times$ & 1.630 & $1.01\times$ \\
\midrule
{\color{famMaple}$\bigstar$} & Maple 20B-A1B & 40.67 & 1.593 & 1.609 & 2.016 & $1.25\times$ & 1.641 & $1.02\times$ \\
\midrule
\multirow{4}{*}{{\color{famBitCPM}$\blacktriangleleft$}}
  & BitCPM-CANN 0.5B         & 37.67 & 1.623 & 1.636 & 2.012 & $1.23\times$ & 1.637 & $1.00\times$ \\
  & BitCPM-CANN 1B           & 38.39 & 1.616 & 1.623 & 2.006 & $1.24\times$ & 1.631 & $1.01\times$ \\
  & BitCPM-CANN 3B           & 38.06 & 1.619 & 1.624 & 2.005 & $1.23\times$ & 1.630 & $1.00\times$ \\
  & BitCPM-CANN 8B           & 39.30 & 1.607 & 1.610 & 2.003 & $1.24\times$ & 1.628 & $1.01\times$ \\
\bottomrule
\end{tabular}
\end{table}

We measure the actual symbol distribution of 29 state-of-the-art (SOTA) ternary LLM models and find that zeros account for up to $51.5\%$ of all weights; Table~\ref{tab:bits} lists the measured zero density of every model. More specifically we benchmarked seven ternary model families:
i) \emph{BitNet}~\cite{bitnet158,bitnet2b4t}, the 2B-parameter model trained from scratch on 4T tokens that established the $1.58$-bit reference point;
ii) \emph{Bonsai}~\cite{bonsai}, four dense checkpoints from $1.7$B to $27$B;
iii) \emph{CAT-Q}~\cite{catq}, five post-training quantizations of Qwen3 from $1.7$B to $235$B, including two mixture-of-experts models;
iv) \emph{ParetoQ}~\cite{paretoq}, five small checkpoints ($125$M--$1.5$B) from a study of low-bit quantization-aware training;
v) \emph{TriLM}~\cite{trilm}, the nine-model Spectra suite ($99$M--$3.9$B) pretrained in ternary;
vi) \emph{Maple}~\cite{maple}, a $20$B-A1B ternary mixture-of-experts reasoning model; and
vii) \emph{BitCPM-CANN}~\cite{bitcpm}, four ternary checkpoints ($0.5$B--$8$B) trained from scratch. Motivated by the finding that the zero density of many ternary models is significantly higher than the density of either of the other two codes $\{-1,+1\}$, we introduce BITCOS (\textbf{\underline{BIT}}map and \textbf{\underline{CO}}mpacted \textbf{\underline{S}}igns): a simple distribution-adaptive layout for ternary tensors. Given a zero density $z$, BITCOS spends one presence bit on every weight entry and one sign bit only on the non-zero weights, thus the layout effectively achieves $2-z$ bits per weight. Such a layout improves the effective bit-width compared to the deployed five-trit packing once $z > 0.375$ and is strictly more efficient than the widely-adopted 2-bit packing for all zero densities. Figure~\ref{fig:models} places every model of Table~\ref{tab:bits} at its measured zero density, on whichever of the two rates it meets first: the sloped $2-z$ line when the BITCOS sparse layout is more efficient, or the $1.625$ plateau when the five-trit fixed-rate packing has lower bit-width. Models with zero density $z$ above $0.375$ benefit from the BITCOS layout (green plot area), as such 26 out of the 29 models achieve lower effective bit-width with BITCOS compared to the five-trit packing layout.

However, the effective bit-width is only one aspect of the overall inference. The actual performance of ternary models during inference also depends on how efficiently the packed weights can be unpacked and utilized in matrix-vector multiplication kernels. The decode phase of LLM inference with a small batch-size is bandwidth-bound, so the time per token tracks the bit-width of the weight datatype~\cite{georganas2025}, which is precisely the regime that on-device and agentic deployments exercise~\cite{slmagentic}. BITCOS not only reduces the storage cost but also is amenable to efficient unpacking and computation on modern CPUs and Intel GPUs. Measured end to end over 7 ternary LLM checkpoints and five platforms, decode throughput improves by up to $1.18\times$ on a $64$-core server CPU, $1.15\times$ on a $24$-core client CPU, $1.27\times$ on a discrete Xe2 GPU and $1.22\times$ on an integrated Xe2 GPU. The performance gain however is not universal: on an 8-core, bandwidth-rich client platform with enough bandwidth per core to leave the unpack sequence exposed, the smaller payload does not translate into performance benefits. Therefore, we develop a simple two-term roofline model to assess the limitations of the BITCOS-based kernels.


This paper makes the following contributions:
\begin{enumerate}
  \item A novel ternary sparse layout (which we name BITCOS) comprising of a presence bitmap plus a compacted sign vector, that costs $2-z$ bits per weight for a zero density $z$ and is amenable to efficient unpacking on modern CPUs and GPUs.
  \item Optimized instruction sequences to unpack the proposed BITCOS layout and perform matrix multiplication operations efficiently on modern x86 CPUs (client CPUs with performance and efficiency cores, and server CPUs with high core counts) and Intel Xe2 GPUs (both integrated and discrete GPUs).
  \item A roofline model to assess the efficacy and the limitations of our CPU microkernels.
  \item Performance evaluation of the proposed layout with microbenchmarks and end-to-end LLM inference over seven ternary LLM checkpoints on modern server/client CPUs and integrated/discrete GPUs, illustrating decode throughput improvements by up to $1.18\times$ on CPUs and $1.27\times$ on GPUs.
\end{enumerate}

\section{The BITCOS Layout: \underline{BIT}map + \underline{CO}mpacted \underline{S}igns}
\label{sec:format}
\subsection{Layout definition and storage cost}
\begin{figure}[t]
\centering
\includegraphics[width=\columnwidth]{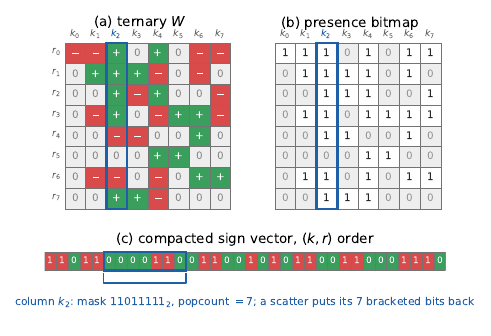}
\caption{The BITCOS layout on an 8$\times$8 tensor. Each column of BITCOS is one
mask whose bit $r$ records whether row $r$ is non-zero, so a decoder
reads presence for a whole block in a single load. The compacted sign vector
carries one bit per non-zero only, in the same $(k,r)$ order, so the $j$-th sign
bit belongs to the $j$-th set bit of the bitmap. Zeros consume a bitmap bit and nothing else, which is why the effective rate is $2-z$. In this example, 30 out of the 64 weights are zero ($z=0.469$), so the tensor costs 64 presence plus 34 sign bits, or $1.531$ bits per weight (which is below $\log_2 3$).}
\label{fig:layout}
\end{figure}

We propose the BITCOS layout that leverages the inherent \emph{unstructured} sparsity of ternary weights and stores a ternary tensor as two pieces:
\begin{enumerate}
  \item a \emph{presence bitmap} of one bit per weight, set where the weight is
        non-zero; and
  \item a \emph{sign vector} of one bit per \emph{non-zero} weight, in tensor
        order.
\end{enumerate}
Figure~\ref{fig:layout} shows both pieces on a small 8$\times$8 tensor.
The bitmap has the full tensor length; the sign vector is compacted to the
population count of the bitmap. Assuming a zero density of $z$, the cost per weight is therefore:
\begin{equation}
  B(z) \;=\; 1 + (1-z) \;=\; 2 - z \quad\text{bits}
  \label{eq:bitmap}
\end{equation}

The cost per weight falls linearly in $z$, so the BITCOS format yields substantial benefits over 2-bit or five-trit packing at zero-heavy distributions. Table~\ref{tab:bits} compares the per-weight cost of different formats over the 29 measured checkpoints. Column ``\% 0'' is the measured zero density $z$, column ``Symbols'' counts the
ternary codes alone and the ``$+$scale'' columns add the measured 16-bit scale overhead of the corresponding model. The ``red.'' columns report the size reduction of BITCOS over the 2-bit and the five-trit per byte packing respectively; a value greater than $1$ means BITCOS stores
the model more compactly than that format. We conclude that BITCOS improves the effective bit-width compared to the deployed five-trit packing once $z > 0.375$ (26 out of 29 ternary LLM models) and is strictly more efficient than the widely-adopted 2-bit packing for all zero densities. 

\subsection{BITCOS unpack sequence with x86 AVX-512 instructions}
\label{sec:decode}

The BITCOS layout reconstructs 16-bit
weights in a short mask-driven sequence assuming the target ISA supports masks. In Figure~\ref{fig:asm512} we illustrate such an exemplary sequence with AVX-512 instructions. Two vectors are prepared once per group of 128 reduction elements and then
reused for every one of its iterations: \texttt{zmm3} holding the 32 fp16 group scales,
one per row, and \texttt{zmm4} holding a copy with the sign bit set, obtained by a single
\texttt{vporq} against a broadcast \texttt{0x8000}. Because the sign of an IEEE
half lives in the most significant bit and the group scales are non-negative by
construction, that OR is an exact negation, so
\texttt{zmm3} and \texttt{zmm4} hold $+s$ and $-s$ respectively for each of the 32 rows.
The per-iteration work is then:
\begin{enumerate}
  \item \textbf{Load the presence mask.} Read 32 bitmap bits into a general
        register and then into \texttt{k1}.
  \item \textbf{Place the signs.} Take the next 32 bits of the sign stream and
        scatter them via \texttt{pdep} into the bit positions that
        \texttt{k1} marks as present. The sign vector is stored compacted, so
        its $j$-th bit belongs to the $j$-th non-zero of the group's bitmap. 
        \texttt{pdep} is exactly the scatter that undoes the compaction (see Figure~\ref{fig:pdep}).
  \item \textbf{Select.} A zero-masked move of \texttt{zmm3} under \texttt{k1}
        puts $+s$ at every present lane and exact $+0$ elsewhere; a merge-masked
        move of \texttt{zmm4} under the deposited mask overwrites the negative
        ones.
\end{enumerate}

Figure~\ref{fig:pdep} illustrates what that deposit does: \texttt{pdep} takes the
low bits of its source in order and drops them at the positions the mask
selects, leaving every unselected position zero. In the AVX-512 assembly, \texttt{r13} walks the bitmap, \texttt{r9} is the base of the sign stream,
\texttt{r8} the running bit position within it, \texttt{r14} walks the
activations, and \texttt{zmm2} is one accumulator for the fused multiply-add (FMA) operation. 
In total we get 17 instructions, of which the weight unpacking portion corresponds to 3 instructions: one \texttt{pdep} and 2 masked moves.  The remaining 14 are not unpacking: one loads the bitmap word, 5 compute the data-dependent address
of the sign window, one \texttt{shrx} performs the unaligned 64-bit read and its
alignment in a single operation, 2 are software prefetches, 2 advance the
bit position by the population count, 2 move masks, and the last is the fused multiply-add (FMA), which absorbs the activation broadcast as an embedded operand.

\begin{figure}[t]
\begin{tcolorbox}[colback=black!3,boxrule=0.4pt,left=4pt,right=4pt,top=3pt,bottom=3pt]
\ttfamily\scriptsize
\setlength{\tabcolsep}{0pt}
\begin{tabular}{@{}l@{\hspace{0.7em}}l@{}}
\multicolumn{2}{@{}l@{}}{\textrm{\emph{once per group of 128:}}} \\
vmovdqu64 & (\%r10,\%rax,2),\%zmm3 \; ; 32 scales \\
vporq     & \%zmm0,\%zmm3,\%zmm4 \; ; negated copy \\[2pt]
\multicolumn{2}{@{}l@{}}{\textrm{\emph{per 32 weights:}}} \\
mov       & -0x404(\%r13,\%rax,8),\%r15d \; ; bitmap[k] \\
mov       & \%r8d,\%ecx \; ; copy bitpos \\
mov       & \%r8,\%rbx \\
shr       & \$0x3,\%rbx \; ; byte offset \\
and       & \%rdx,\%rbx \; ; to 32-bit word \\
and       & \$0x1f,\%cl \; ; shift count \\
shrx      & \%rcx,(\%r9,\%rbx,1),\%rdi \; ; sign window \\
prefetcht0 & -0x4(\%r13,\%rax,8) \; ; bitmap \\
popcnt    & \%r15d,\%ecx \\
prefetcht0 & 0x200(\%r9,\%rbx,1) \; ; signs \\
add       & \%r8,\%rcx \; ; bitpos += popcnt \\
pdep      & \%r15d,\%edi,\%edi \; ; to lane order \\
kmovd     & \%r15d,\%k1 \; ; presence mask \\
vmovdqu16 & \%zmm3,\%zmm5\{\%k1\}\{z\} \; ; $+s$, else $+0$ \\
kmovd     & \%edi,\%k1 \; ; sign mask \\
vmovdqu16 & \%zmm4,\%zmm5\{\%k1\} \; ; $-s$ if negative \\
vfmadd231ph & -0x2(\%r14,\%rax,4)\{1to32\},\%zmm5,\%zmm2 \\
\end{tabular}
\end{tcolorbox}
\caption{AVX-512 instruction sequence to unpack the BITCOS layout for one 32-row block.}
\label{fig:asm512}
\end{figure}

\begin{figure}[t]
\begin{tcolorbox}[colback=black!3,boxrule=0.4pt,left=4pt,right=4pt,top=3pt,bottom=3pt]
\ttfamily\scriptsize
\textrm{\emph{dst := \_pdep\_u32(a, mask)}} \\[2pt]
dst := 0; k := 0 \\
FOR m := 0 TO 31 \\
\hspace*{1.2em} IF mask[m] == 1 THEN \\
\hspace*{2.4em} dst[m] := a[k]; k := k + 1 \\
\hspace*{1.2em} FI \\
ENDFOR \\[5pt]
\textrm{\emph{example, bit 0 leftmost, 8 lanes shown:}} \\[3pt]
\setlength{\tabcolsep}{0pt}
\begin{tabular}{@{}l@{\hspace{0.8em}}*{8}{c@{\hspace{0.55em}}}@{\hspace{0.5em}}l@{}}
mask & 1&0&1&1&0&0&1&0 & \textrm{\scriptsize presence: 4 non-zeros} \\
a    & 1&0&1&1& \textrm{\tiny --}& \textrm{\tiny --}& \textrm{\tiny --}& \textrm{\tiny --} & \textrm{\scriptsize signs, compacted} \\
dst  & 1&0&0&1&0&0&1&0 & \textrm{\scriptsize signs, lane order} \\
\end{tabular}
\end{tcolorbox}
\caption{\texttt{pdep} parallel bit deposit. In this example, only the low four bits of \texttt{a} are consumed, one per set bit of
the mask. The sign stream stores one bit per non-zero weight, so undoing
that compaction means scattering those bits onto the set positions of the
presence mask, which is the instruction's definition.}
\label{fig:pdep}
\end{figure}

\subsection{BITCOS unpack sequence with x86 AVX2 instructions}
\label{sec:decode-avx2}

The sequence of Section~\ref{sec:decode} depends on two AVX-512 facilities:
mask registers, which make a 32-bit presence pattern directly usable as a
predicate, and FP16 compute instructions. The client CPUs we target support neither, thus we implemented a second kernel targeting the AVX2 ISA with AVX-VNNI-INT8 compute capabilities, which reconstructs the ternary weights
$\{-1,0,+1\}$ as \texttt{int8} and contracts against \texttt{int8} activations (see Figure~\ref{fig:asmavx2}).

For each of the 32 rows, let $p_i\in\{0,1\}$ indicate that the weight is
present, and let $n_i\in\{0,1\}$ indicate that the present weight is negative. The bitmap supplies the $p_i$ bits. After \texttt{pdep}
returns the compact signs to their row positions, its result supplies the
$n_i$ bits. Since AVX2 has no mask registers, we
materialize each presence and sign vector as byte masks
$P_i=-p_i$ and $N_i=-n_i$: a true lane is \texttt{0xFF} ($-1$ as a signed
byte), and a false lane is zero. The desired ternary byte is then
\begin{equation}
  w_i = p_i - 2n_i .
  \label{eq:avx2ternary}
\end{equation}
In the example of Figure~\ref{fig:layout}, the column $k_2$ holds the ternary values
$(+1,+1,+1,+1,-1,0,-1,+1)$ for rows $i = 0\ldots7$, so the bitmap word for that
column is $\mathtt{0xDF}$ and the two predicates are
\begin{equation}
  p = (1,1,1,1,1,0,1,1), \qquad n = (0,0,0,0,1,0,1,0).
  \label{eq:avx2example}
\end{equation}
Equation~\eqref{eq:avx2ternary} returns the unpacked column exactly: rows $0$--$3$ and
$7$ give $1-0=+1$, rows $4$ and $6$ give $1-2=-1$, and the absent row $5$ gives
$0-0=0$. Materialized as bytes these are
$P=(\mathtt{FF},\mathtt{FF},\mathtt{FF},\mathtt{FF},\mathtt{FF},\mathtt{00},\mathtt{FF},\mathtt{FF})$
and, after the mask with \texttt{0xFE},
$T=(\mathtt{00},\mathtt{00},\mathtt{00},\mathtt{00},\mathtt{FE},\mathtt{00},\mathtt{FE},\mathtt{00})$,
so \texttt{vpsubb} leaves
$(\mathtt{01},\mathtt{01},\mathtt{01},\mathtt{01},\mathtt{FF},\mathtt{00},\mathtt{FF},\mathtt{01})$,
which is the original column as \texttt{int8}. Note that $n$ is not what the
format stores. Only the $\operatorname{popcount}(p) = 7$ sign bits
$(0,0,0,0,1,1,0)$ are stored, so the sign of row $6$ is the sixth stored bit rather
than the seventh. Recovering the $n$ of \eqref{eq:avx2example} from those seven
bits is exactly the \texttt{pdep} of Figure~\ref{fig:asmavx2}.

\begin{figure}[!t]
\begin{tcolorbox}[colback=black!3,boxrule=0.4pt,left=2pt,right=2pt,top=2pt,bottom=2pt]
\ttfamily\scriptsize
\setlength{\tabcolsep}{0pt}
\renewcommand{\arraystretch}{0.96}
\begin{tabular}{@{}l@{\hspace{0.5em}}l@{}}
\multicolumn{2}{@{}l@{}}{\textrm{\emph{once per quad of reduction elements:}}} \\
vbroadcastss & (\%r9,\%r10,4),\%ymm8 \; ; activation quad \\[1pt]
\multicolumn{2}{@{}l@{}}{\textrm{\emph{per 32 weights:}}} \\
mov      & (\%rsi,\%rcx,4),\%ebx \; ; bitmap word \\
popcnt   & \%ebx,\%edx \\
add      & \%r14,\%rdx \; ; bitpos += popcnt \\
shr      & \$0x3,\%r14 \\
and      & \%rbp,\%r14 \; ; to 32-bit word \\
and      & \$0x1f,\%r15b \; ; shift count \\
shrx     & \%r15,(\%rdi,\%r14,1),\%r14 \; ; sign window \\
pdep     & \%ebx,\%r14d,\%r14d \; ; to lane order \\
vmovd    & \%ebx,\%xmm9 \\
vpbroadcastd & \%xmm9,\%ymm9 \\
vpshufb  & \%ymm11,\%ymm9,\%ymm9 \; ; spread bitmap bytes \\
vpand    & \%ymm12,\%ymm9,\%ymm9 \; ; isolate row bits \\
vpcmpeqb & \%ymm12,\%ymm9,\%ymm9 \; ; $P_i=-p_i$ \\
vmovd    & \%r14d,\%xmm10 \\
vpbroadcastd & \%xmm10,\%ymm10 \\
vpshufb  & \%ymm11,\%ymm10,\%ymm10 \\
vpand    & \%ymm12,\%ymm10,\%ymm10 \\
vpcmpeqb & \%ymm12,\%ymm10,\%ymm10 \; ; $N_i=-n_i$ \\
vpand    & \%ymm13,\%ymm10,\%ymm10 \; ; $T_i=-2n_i$ \\
vpsubb   & \%ymm9,\%ymm10,\%ymm9 \; ; $w_i=T_i-P_i$ \\
vpdpbssd & \%ymm8,\%ymm9,\%ymm7 \\
\end{tabular}
\end{tcolorbox}
\caption{AVX2 instruction sequence to unpack the BITCOS layout for one 32-row block.}
\label{fig:asmavx2}
\end{figure}

\begin{figure}[!t]
\begin{tcolorbox}[colback=black!3,boxrule=0.4pt,left=2pt,right=2pt,top=2pt,bottom=2pt]
\scriptsize
\raggedright
\setlength{\parindent}{0pt}
\setlength{\parskip}{0pt}
\def\sect#1{\vspace{1.5pt}\textrm{\emph{#1}}\par}
\def\ins#1#2{\setbox0=\hbox{\texttt{#1}\hspace{0.5em}}\hangindent=\wd0\hangafter=1\noindent\unhbox0{\color[HTML]{1B6B3A}\ttfamily// #2}\par}
\sect{per two 32-row blocks:}
\ins{load.ugm.d32x3 r21, [p+rank/32]}{3 sign words per lane, returned in r21--r23}
\sect{per 32-row block, i.e.\ eight groups:}
\ins{load\_block2d.ugm bmp, [...]}{one contiguous bitmap word for each of 16 output columns}
\ins{bfn w0, r21, r22, sel}{mux: select the gathered word containing the window's low bits}
\ins{bfn w1, r22, r23, sel}{mux: the following word, for bits that cross the boundary}
\ins{and.eq f0, b, rank, 31}{$b=rank\bmod32$}
\ins{shr lo, w0, b}{align low bits at current bit rank}
\ins{shl hi, w1, (-rank)\&31}{complementary count $(32-b)\bmod32$}
\ins{(!f0) or signs, lo, hi}{at $b=0$ the high word contributes nothing}
\ins{cbit nsg, bmp}{\# of nonzeros in 32-row bitmap block}
\ins{add rank, rank, nsg}{advance the block-level sign rank}
\sect{per four-row group:}
\ins{shr nib, bmp, imm\_g}{extract group's 4 presence bits}
\ins{and r68, nib, idxmask}{presence nibble in LUT-key field}
\ins{cbit r68, r68}{\# of sign bits consumed by this group}
\ins{shl r66, signs, 3}{shift sign window in key field}
\ins{and r66, r66, signmask}{retain next 4 compact sign bits}
\ins{bfn r67, nib, idxmask, r66}{merge: presence and sign fields into one byte offset}
\ins{add r67, r67, lut\_base}{add runtime SLM base of 2KB LUT}
\ins{load.slm.d32x2 r74, [r67]}{16 lookups, each lane receives 4 fp16 ternary codes}
\ins{shr signs, signs, r68}{consume only the sign bits used by present rows}
\ins{mul r54, scale, r74}{multiply ternary codes with scales}
\sect{once per four groups:}
\ins{dpas.8x1 acc, acc, r54, act}{fp16 XMX contraction}
\end{tcolorbox}
\caption{Xe2 instruction sequence to unpack the BITCOS format. Register names are shortened and independent operations are
grouped by function rather than scheduler order. Address setup, predicate
formation, and register repacking are omitted. The 16 SIMD
lanes correspond to different output columns, so one four-row group produces
$4\times16$ weights. Section~\ref{sec:decode-xe2} walks through the sequence.}
\label{fig:xe2asm}
\end{figure}

This alternative contraction algorithm has two implications. First, arithmetic moves from fp16 to integer: activations
are quantized to \texttt{int8} per group of 128, the dot product accumulates in
\texttt{int32}, and the group scales are applied once per group instead of being
blended into the weights. The weights are therefore reconstructed as the values
$\pm1$ rather than $\pm s$, and the layout is stored in VNNI4 order so that each
\texttt{vpdpbssd} (AVX-VNNI-INT8 compute) consumes four consecutive reduction elements for each of eight
rows. Second, and more consequentially, the absence of mask registers means the
presence pattern must be materialized as one \emph{byte} per lane before it can
select anything. That expansion consists of a broadcast, an in-lane shuffle and a compare against the bit-select constant, and it costs 5
instructions where AVX-512 spends merely 1 mask move instruction \texttt{kmovd}. Once those byte masks exist, the last two
instructions implement Eq.~\eqref{eq:avx2ternary}: masking $N_i$ with
\texttt{0xFE} forms $T_i=-2n_i$, and the \texttt{vpsubb P,T,W} computes $W_i=T_i-P_i=p_i-2n_i$. Thus a positive
present lane becomes $+1$, a negative present lane becomes $-1$, and an absent
lane remains zero.

\subsection{BITCOS unpack sequence for Intel Xe2 GPUs}
\label{sec:decode-xe2}

Both x86 sequences rely on \texttt{pdep} to scatter compact signs back to the rows marked present. Xe2 has no corresponding
instruction, so the GPU kernel replaces that scatter with a small lookup table
in shared local memory (SLM). We implement the kernel with the XeTLA
templates~\cite{xetla}, and use the conventional SYCL terminology of workgroups
and subgroups throughout~\cite{sycl}. Figure~\ref{fig:xe2asm} illustrates the assembly sequence for the Xe2 kernel.

In the assembly sequence of Figure~\ref{fig:xe2asm}, one
\texttt{load.ugm.d32x3} fetches the three consecutive sign words sufficient for
two 32-row blocks. For each block, the contiguous bitmap load \texttt{load\_block2d.ugm bmp} supplies one
32-bit presence word per output column. Two \texttt{bfn}
instructions, the three-input bitwise Boolean operation of Xe2, select the
adjacent low/high sign words, and an \texttt{and}, two shifts and an \texttt{or}
align them at the current bit rank. The \texttt{and} reduces the rank to the
in-word offset $b$, the \texttt{shr} aligns the low word by $b$ and the
\texttt{shl} brings in the $b$ high bits that cross the word boundary. That last
shift needs a count of $32-b$, so the complementary count
is taken modulo 32 and the \texttt{or} is predicated off in the one case the wrap
gets wrong, $b=0$, where the high word contributes nothing. The block-wide \texttt{cbit} advances that
rank by the number of nonzeros in all 32 rows. Each block is then decoded as eight four-row groups. A group extracts one
presence nibble (4 bits), combines it with the next 4 bits of the aligned sign
window, and uses the resulting byte offset for one SIMD16
\texttt{load.slm.d32x2}. The lookup returns four fp16 ternary codes per lane.
The group-level \texttt{cbit} advances the sign window by only the bits actually
consumed, an fp16 multiply applies the group scale, and one DPAS contraction is
issued after four groups have supplied 16 reduction rows. The sequence touches
two memories, \texttt{load.ugm} for the presence bitmap and sign vector in global
memory and \texttt{load.slm} for the table in SLM.

\begin{figure}[!t]
\centering
\includegraphics[width=\columnwidth]{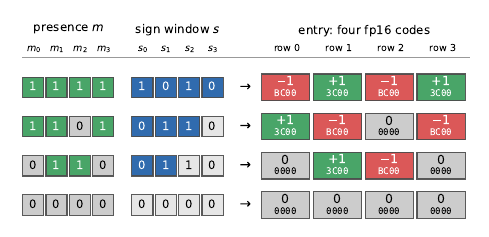}
\caption{Representative entries of the $256$-entry lookup table. The key is the
presence nibble $m$ together with a four-bit window $s$ of the compact sign
stream, and the entry holds four fp16 constants, shown with their bit patterns. The second and third
rows are the first two
groups worked through in Figure~\ref{fig:xe2lut}, with keys $\mathtt{0xB6}$ and
$\mathtt{0x66}$.}
\label{fig:xe2entries}
\end{figure}

The lookup table in SLM is indexed by an 8-bit key formed
from two nibbles: the four presence bits $m$ of the bitmap for rows $4g\ldots
4g{+}3$, and the next four bits $s$ of that column's compact sign stream. Entry
$(m,s)$ holds four fp16 constants $c_0\ldots c_3$, one per row, where
\begin{equation}
  c_i =
  \begin{cases}
    0 & \text{if } m_i = 0,\\
    +1 & \text{if } m_i = 1 \text{ and } s_{r_i} = 0,\\
    -1 & \text{if } m_i = 1 \text{ and } s_{r_i} = 1,
  \end{cases}
  \qquad r_i = \textstyle\sum_{j<i} m_j ,
  \label{eq:lutentry}
\end{equation}
and $r_i$ is the rank of row $i$ among the present rows of the nibble. The table is a precomputed, four-bit-wide \texttt{pdep}
composed with the map from sign bit to ternary code; Figure~\ref{fig:xe2entries} lists representative entries.

\begin{figure}[!h]
\centering
\includegraphics[width=\columnwidth]{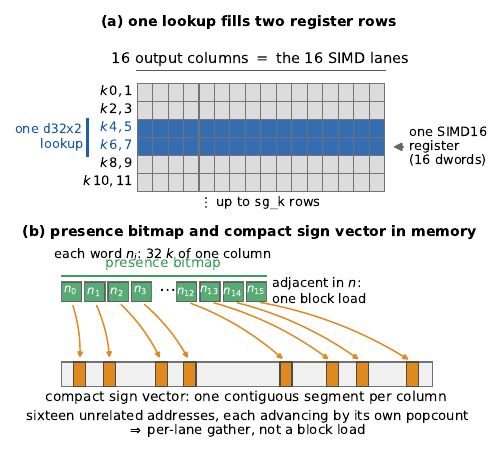}
\caption{Xe2 kernel mapping of a weight tile onto SIMD lanes. (a) Each grid column is one of the 16 output columns and is held
by one SIMD lane, and each grid row is one register row. VNNI2 order puts a
$k$-pair in every dword. The blue-shaded band is the output of a single
\texttt{load.slm.d32x2}, which returns two dwords per lane and therefore fills
2 register rows, i.e.\ 4 reduction rows, in one lookup. (b) The presence
bitmap and the compact sign vector in memory, for the
16 columns one subgroup owns: the bitmap words of adjacent columns are
adjacent, whereas each column enters the sign vector at its own offset.}
\label{fig:xe2lanes}
\end{figure}
\begin{figure}[!h]
\centering
\includegraphics[width=\columnwidth]{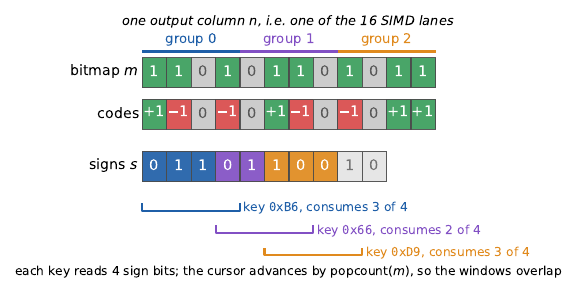}
\caption{Three consecutive four-row lookups on one column. Each group forms an
8-bit key from its presence nibble and a four-bit window of the compact sign
stream. The bracket under each window gives the resulting table index/key. The sign-stream consumption is disjoint, but because the
cursor advances by $\operatorname{popcount}(m)$, the read
windows overlap. The figure follows a single
column, that is one SIMD lane. The kernel runs 16 of these in parallel, each with its own presence word, its own entry point into the sign stream and its own
cursor.}
\label{fig:xe2lut}
\end{figure}

Because a column's sign bits are compacted, reading them requires tracking a
per-column read position, which we call that column's \emph{cursor}: the number
of sign bits the groups above it have already consumed, equivalently the rank of
its next non-zero row among the non-zeros seen so far. The cursor is an offset into a variable-rate stream, and it is the only piece of
per-column state whose value the bitmap alone does not give away. The sign field of the key always takes four bits, because four is the most a
four-row group can need, but only $\operatorname{popcount}(m)$ of them are
consumed; that population count is a single \texttt{cbit} in
Figure~\ref{fig:xe2asm}. The cursor therefore advances by
$\operatorname{popcount}(m)$ and the
next group re-reads whatever this one left behind, so successive read windows
\emph{overlap} even though the bits they consume are disjoint, as
Figure~\ref{fig:xe2lut} draws for three consecutive groups. This is why the
key is formed by a shift and a mask of a running window rather than by indexing
the stream. The kernel keeps the window in a register and shifts it right by
$\operatorname{popcount}(m)$ after each group (\texttt{shr signs} of
Figure~\ref{fig:xe2asm}). The gather address is maintained more coarsely, where the kernel takes one
\texttt{cbit} of the entire 32-bit presence word per block and adds that to the column rank. The two population counts \texttt{cbit} in Figure~\ref{fig:xe2asm} therefore
serve different purposes and are not redundant, i.e.\ the per-group one only drives
the window shift, and the per-block one only drives the gather address. Keeping
them apart is what keeps the loop-carried chain short, since the address
depends on one count per thirty-two rows rather than on a chain of eight. The
per-group count also depends only on the bitmap, so it can issue before the SLM
lookup returns.
Because $m$ and $s$ are
four bits each, the table has $2^8 = 256$ entries of four fp16 values, so its size is
$2$\,KB in total. At kernel entry, the workgroup's subgroups cooperatively initialize
disjoint table entries in one SLM-resident LUT. The group size of four rows is not arbitrary. Four presence bits
and four sign bits give a table small enough to sit in SLM, whereas an eight-row group would
need $2^{16}$ entries. Four is also compatible with the XMX/DPAS operand layout. The fp16 DPAS consumes the weight tensor in VNNI2 order, with reduction rows $2m$
and $2m{+}1$ packed into one dword, so four \emph{consecutive} rows are
precisely two VNNI2 dwords. The entry is therefore stored in the order the tile
needs, a single \texttt{d32x2} lookup returns both dwords, the
\texttt{load.slm.d32x2} of Figure~\ref{fig:xe2asm}, and they are written
into the unpacked tile with no shuffle or transpose, thus meeting the VNNI2
requirement without any extra instructions.

Figure~\ref{fig:xe2lanes}(a) makes the mapping concrete. A subgroup builds a
tile of sixteen weight columns by \texttt{sg\_k} reduction rows, and because a
dword holds two consecutive $k$ of one column, one SIMD16 register is exactly
one $k$-pair across all sixteen columns. One lookup therefore fills two
adjacent register rows, that is four reduction rows of sixteen columns, or
$64$ weights per message. The sixteen lanes are sixteen \emph{output columns}, so the sequence is
vectorized along $n$ while $k$ is walked serially by the loop. A column's cursor depends on the population counts of all its preceding
groups, so neighboring lanes drift apart as they advance.
Figure~\ref{fig:xe2lanes}(b) shows the presence bitmap and the compact sign vector in memory. The presence
bitmap is stored as $\lceil K/32\rceil\times N$ words, each folding $32$
reduction rows of one column, and the words of adjacent columns are themselves
adjacent, so sixteen lanes read them with a single block load, the
\texttt{load\_block2d.ugm} of Figure~\ref{fig:xe2asm}. The sixteen sign
cursors, by contrast, are unrelated addresses, so the sign words must be
gathered per lane (the gather \texttt{load.ugm.d32x3}). One 32-row window needs two adjacent words,
and the next block starts at most one word later, so the union of both windows
is exactly three words and a single \texttt{d32x3} serves two blocks rather than
one. A fourth word is never consumed: the cursor can sit at most $31$ bits into
a word and two blocks consume at most $32$ sign bits each, so the span reaches
bit $31+64-1 = 94$ at worst, still inside the third word.
In group~$0$ of Figure~\ref{fig:xe2lut} the bitmap nibble $m = 1011_2$ marks rows $0$, $1$ and $3$ as
non-zero and row~$2$ as zero, and the next four compact sign bits are
$s = 0110_2$, so the key is table index $\mathtt{0xB6}$. Row~$0$ takes sign
$s_0 = 0$ and row~$1$ takes $s_1 = 1$, but row~$2$ is absent and consumes
nothing, so row~$3$ takes $s_2 = 1$ rather than $s_3$. The entry is therefore
$(+1,-1,0,-1)$, and the cursor advances by
$\operatorname{popcount}(m) = 3$, and the following groups advance it
by $2$ and $3$.

\section{Experimental results}
\label{sec:experiments}

\subsection{Experimental platforms}
\label{sec:platforms}

We use three x86 CPU platforms with different core counts, core types and
memory bandwidth:

\begin{itemize}
\item One socket of an Intel Xeon Platinum 8592+ CPU (referred to as EMR) with
$64$ cores. It has DDR5@4400 MT/s memory and a measured streaming
read bandwidth of $\sim245$\,GB/s. It supports Advanced Matrix Extensions (AMX) and AVX-512, including AVX-512-FP16.
\item An Intel Core Ultra 9 285K CPU (referred to as ARL) with $24$ cores ($8$ performance cores and $16$ efficiency cores). It has
dual-channel DDR5 memory and a measured read bandwidth of $\sim98$\,GB/s. It
supports AVX-VNNI-INT8 but not AVX-512.
\item An Intel Core Ultra 7 258V CPU (referred to as LNL CPU) with $8$ cores ($4$ performance and $4$ efficiency cores). It has
$32$\,GB of LPDDR5X memory and a measured read bandwidth of $\sim108$\,GB/s. It supports AVX-VNNI-INT8 but not AVX-512.
\end{itemize}

The instruction sets are relevant to the results: EMR runs the AVX-512 kernel of Section~\ref{sec:decode}, and ARL and LNL run the AVX2 kernel of
Section~\ref{sec:decode-avx2}.
For the GPU evaluation we use two Xe2 GPU platforms:

\begin{itemize}
\item The Intel Arc 140V, which is the integrated GPU of the LNL platform above.
It has $8$ Xe2 cores, and it uses the same LPDDR5X memory as the LNL
CPU, and its measured read bandwidth is $\sim108$\,GB/s.
\item An Intel Arc Pro B70 discrete GPU. It has
$32$ Xe2 cores and $32$\,GB of dedicated GDDR6 memory, with a
measured read bandwidth of $\sim500$\,GB/s.
\end{itemize}

\subsection{Results on the CPU platforms}
\label{sec:cpuresults}
We first introduce a roofline model to assess the efficacy and limitations of our CPU kernels, and then present the GEMV microbenchmarks and the end-to-end decode results.

\subsubsection{A roofline model for the BITCOS CPU kernels}
\label{sec:roofline}

\begin{figure}[t]
\centering
\includegraphics[width=\columnwidth]{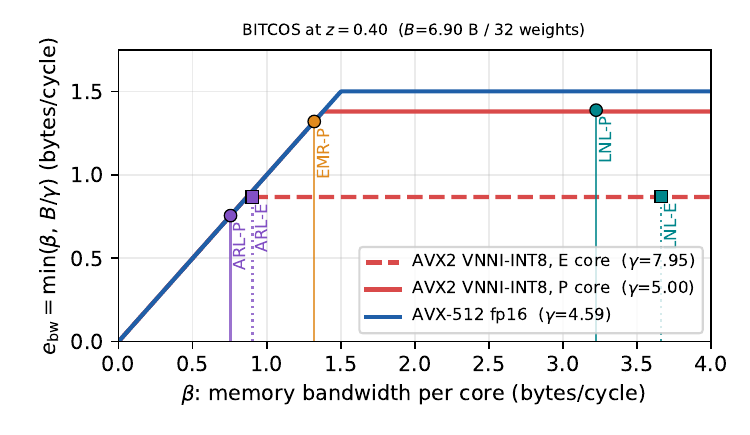}
\caption{BITCOS CPU roofline at zero density $z=0.40$. Each line is
$e_{\mathrm{bw}}=\min(\beta,B/\gamma)$ for one kernel and core type; the flat part is the
instruction ceiling $B/\gamma$ and the diagonal is the memory bandwidth limit. Markers
place the five measured core groups. Emerald Rapids and the Arrow Lake performance cores sit on the diagonal (thus the kernels are bandwidth bound), the Arrow Lake efficiency cores sit at the knee, and Lunar Lake sits on the flat part (thus it is instruction bound).}
\label{fig:roofline}
\end{figure}

\begin{figure*}[t]
\centering
\includegraphics[width=\textwidth]{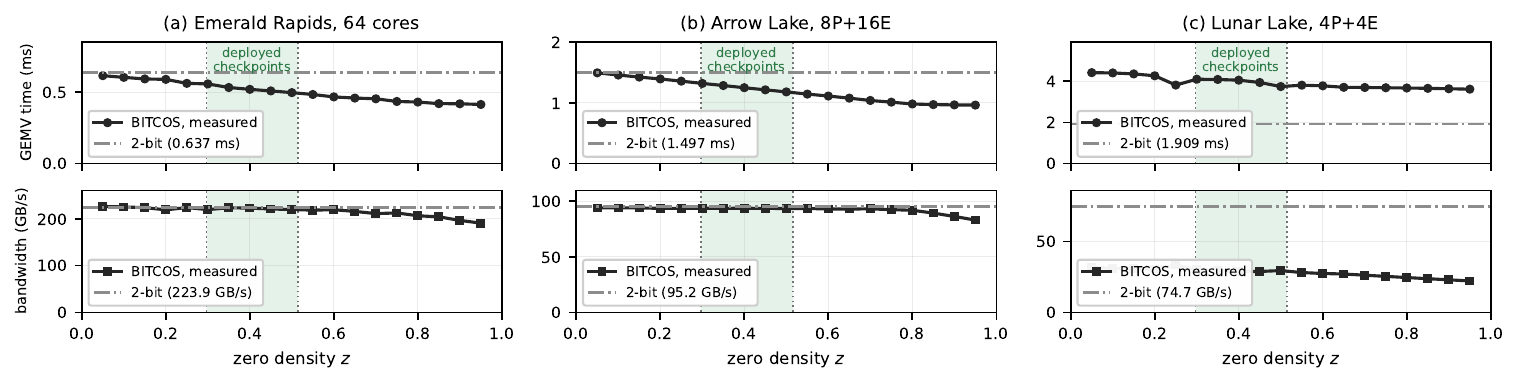}
\caption{Zero-density sweep for a $32k\times16k$ matrix-vector multiplication on: (a) Emerald Rapids, (b) Arrow Lake and (c) Lunar Lake, against the flat LIBXSMM
2-bit reference. The shaded band marks $z\in[0.297,0.515]$, the range spanned
by the deployed checkpoints of Table~\ref{tab:bits}. Top: GEMV time. Bottom:
effective bandwidth.}
\label{fig:cpushmoo}
\end{figure*}

\begin{table}[t]
\centering
\footnotesize
\setlength{\tabcolsep}{4pt}
\caption{BITCOS CPU roofline at zero density $z=0.40$. $\gamma$ is the measured
cost in cycles of one L1-resident microkernel iteration and $\beta$ is the
per-core share of the measured read bandwidth. The bounding/bottleneck term is
set in bold.}
\label{tab:roofcpu}
\begin{tabular}{llrrrrl}
\toprule
& & \textbf{\# of} & & & & \\
\textbf{platform} & \textbf{core type} & \textbf{cores} &
$\gamma$ & $\beta$ & $B/\gamma$ & \textbf{bound by} \\
\midrule
EMR & P & 64 & 4.60 & \textbf{1.321} & 1.502 & memory \\
ARL & P & \phantom{0}8 & 5.00 & \textbf{0.756} & 1.381 & memory \\
ARL & E & 16 & 7.95 & 0.903 & \textbf{0.868} & instructions (knee) \\
LNL & P & \phantom{0}4 & 5.00 & 3.224 & \textbf{1.381} & instructions \\
LNL & E & \phantom{0}4 & 7.95 & 3.663 & \textbf{0.868} & instructions \\
\bottomrule
\end{tabular}
\end{table}

In this section we present a simple two-term bottleneck roofline
model~\cite{roofline}, similar to the one used for the fixed-width ternary kernels in prior work~\cite{georganas2025}. For both the AVX-512 and the AVX2 CPU microkernels (i.e.\ see Figures~\ref{fig:asm512} and \ref{fig:asmavx2}), one iteration of the innermost loop upconverts $32$ ternary weight values. Let $\gamma$ be the number of cycles that iteration costs when every operand is already in L1, and
$\beta$ the share of read bandwidth available to one core, in bytes per cycle.
For these 32 weight entries, an iteration reads (in bytes):
\begin{equation}
B(z) = \underbrace{4}_{\text{bitmap}} + \underbrace{4(1-z)}_{\text{signs}}
     + \underbrace{0.5}_{\text{fp16 scale}} = 8.5 - 4z
\end{equation}
so the time $T$ per iteration and the bandwidth $e_{\mathrm{bw}}$ a core can sustain are:
\begin{equation}
T = \max\!\left(\frac{B(z)}{\beta},\, \gamma\right), \qquad
e_{\mathrm{bw}} = \min\!\left(\beta,\, \frac{B(z)}{\gamma}\right).
\label{eq:roofline}
\end{equation}

The BITCOS-based kernel is memory-bound while $B(z)/\gamma > \beta$ and instruction-bound
otherwise. The fixed-width 2-bit kernels have a constant $B$ while a BITCOS-based kernel does not, so its knee moves with the density of the model at hand since $B(z)$ depends on $z$. For the per core bandwidth $\beta$, we take the per-core share of the streaming read bandwidth of the corresponding platform. We measure $\gamma$ empirically on each platform by running the microkernel loop over an L1-resident block. Table~\ref{tab:roofcpu} illustrates the measured $\gamma$ values for the various core types, and we also report the term $B/\gamma$ for a zero density $z=0.4$ which implies $B=6.9$ bytes read per 32 weight entries. Arrow Lake and Lunar Lake have the same performance and efficiency cores and run the same AVX2 microkernel, so a single pair of measurements for $\gamma$ on performance and efficiency cores serves both platforms. Table~\ref{tab:roofcpu} also evaluates Equaå~\eqref{eq:roofline} for the three CPU platforms, and
Figure~\ref{fig:roofline} depicts the corresponding roofline. Emerald Rapids is memory-bound on all of its cores and Arrow Lake on its performance cores, so the BITCOS-based kernel converts its smaller
payload into savings in execution time; the Arrow Lake efficiency cores sit at the knee, where the two terms are within $4\%$ of each other. Lunar Lake is instruction-bound on both core types: with only eight cores
sharing $108$\,GB/s, each core has $3.2$--$3.7$ bytes per cycle available but
can only consume $0.87$--$1.38$ bytes per cycle, so roughly three quarters of the
bandwidth the platform offers a core is not attainable for this kernel. A bandwidth-rich client platform with a limited number of cores is exactly the case where a cheaper decode (like the 2-bit kernels from prior work~\cite{georganas2025}) beats kernels with smaller payload and more expensive decode (like the BITCOS-based kernel of this work).

\subsubsection{GEMV microbenchmarks on Emerald Rapids}
\label{sec:emr}
To test the efficacy of the BITCOS-based GEMV microkernel we experimented with a large 32768$\times$16384 ternary weight matrix. Weights are replicated to a working set of
at least $4$\,GB so that nothing is served from the last-level cache. We use group size 128, i.e.\ 128 entries along the inner-product dimension share one 16-bit scale, and we vary the zero density $z$. The 2-bit reference GEMV is the production LIBXSMM 2-bit microkernel for CPUs~\cite{georganas2025}. In Figure~\ref{fig:cpushmoo}(a) top panel we illustrate the execution time of the GEMV on EMR, whereas on the bottom panel we show the corresponding effective bandwidth. We observe that the BITCOS format wins at every density, and converts most of its bit-width advantage into execution time savings (see Figure~\ref{fig:cpushmoo}(a) bottom panel, where the effective bandwidth stays constant $\sim225$\,GB/s for $z$ up to $0.6$). This behavior is also validated by our roofline model, where on EMR the BITCOS kernel operates in a bandwidth-bound regime for $z < 0.6$. In these plots we highlight with a green area the zero densities of interest: the deployed ternary models/checkpoints we examined in Table~\ref{tab:bits} exhibit $z\in[0.297,0.515]$. For these zero densities, the observed speedup of the BITCOS-based GEMV over the 2-bit SOTA GEMV is in the range of $1.14$--$1.28\times$. For extreme zero densities (e.g.\ $z=0.95$) we observe that the per-iteration byte count drops, yielding $B/\gamma = 1.02$, and the unpack instruction sequence starts being the bottleneck, thus restricting the effective bandwidth to $190.6$\,GB/s.

\subsubsection{GEMV microbenchmarks on Arrow Lake}
\label{sec:arl}
We repeat the same GEMV benchmark on Arrow Lake (see Figure~\ref{fig:cpushmoo}(b)) and the conclusions are the same as the ones on EMR: the BITCOS-based GEMV wins at every density, and converts most of its bit-width advantage into execution time savings, which is in alignment with our roofline analysis. Over the same band of deployed zero densities, $z\in[0.297,0.515]$, the observed speedup of the BITCOS-based GEMV over the 2-bit SOTA GEMV is in the range of $1.13$--$1.27\times$, and the kernel holds $93.4$--$93.6$\,GB/s of effective bandwidth across that band.

\begin{figure*}[t]
\centering
\includegraphics[width=\textwidth]{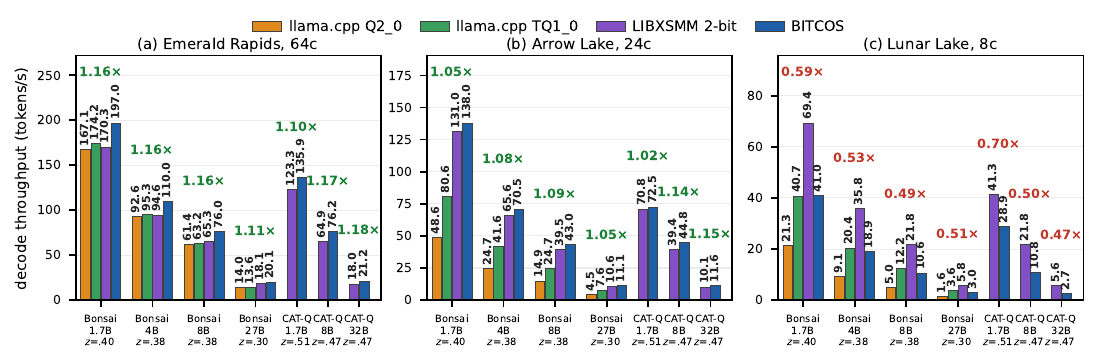}
\caption{Decode throughput in tokens per second on the three CPU platforms,
batch one, over 7 ternary LLMs, with each model's measured
zero density $z$ under its name. The orange and green bars correspond to the Prism ML fork of \texttt{llama.cpp} and appear only for the Bonsai family of models that the fork supports. The purple bar corresponds to the SOTA LIBXSMM 2-bit kernel~\cite{georganas2025} and the blue bar is BITCOS (this work), both inside the vLLM CPU backend. The green number above each cluster of bars is the speedup of BITCOS over the SOTA 2-bit kernel.}
\label{fig:e2ecpu}
\end{figure*}

\subsubsection{GEMV microbenchmarks on Lunar Lake}
\label{sec:lnl}
 Figure~\ref{fig:cpushmoo}(c) illustrates the GEMV benchmark on Lunar Lake CPU, where our roofline model predicts that the BITCOS-based GEMV kernel is instruction-bound on both core types, and as such it is expected to be slower than the SOTA 2-bit GEMV. The measurements in Figure~\ref{fig:cpushmoo}(c) bottom panel confirm that prediction: the BITCOS kernel sustains only $28.9$\,GB/s of effective bandwidth at $z=0.40$. The BITCOS kernel is slower than the 2-bit reference at every density, and its effective bandwidth never exceeds $33.3$\,GB/s against the $74.7$\,GB/s the two-bit kernel sustains. This result confirms our roofline analysis, and it is a cautionary tale for client platforms with a limited number of cores and high memory bandwidth per core: a 2-bit, cheaper decode GEMV kernel beats kernels with smaller payload and more expensive decode.

\subsubsection{End-to-end decode on the CPU platforms}
\label{sec:e2e}
\begin{figure*}[!t]
\centering
\includegraphics[width=\textwidth]{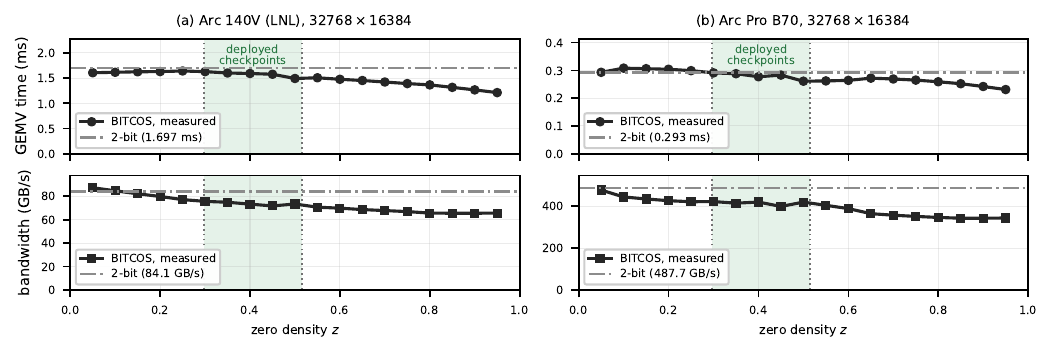}
\caption{Zero-density sweep for a $32k\times16k$ matrix-vector multiplication on (a) the
Arc 140V and (b) the Arc Pro B70, against the flat XeTLA int2
reference, with every point tuned independently. The
shaded band marks $z\in[0.297,0.515]$, the range spanned by the
deployed checkpoints of Table~\ref{tab:bits}. Top: GEMV time. Bottom: effective
bandwidth.}
\label{fig:xe2shmoo}
\end{figure*}

\begin{figure*}[t]
\centering
\includegraphics[width=\textwidth]{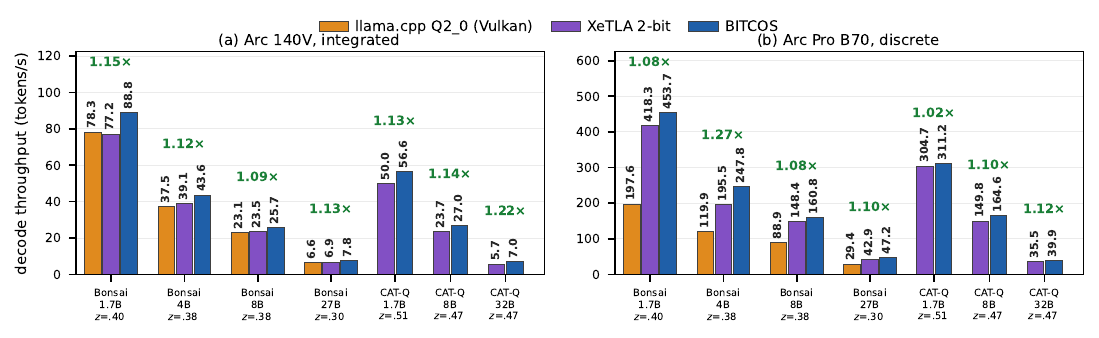}
\caption{Decode throughput in tokens per second on the two Xe2 platforms, batch
one, over 7 ternary LLMs, with each model's measured zero
density $z$ under its name. The orange bar corresponds to the Prism ML fork of
\texttt{llama.cpp} on its Vulkan backend and appears only for the Bonsai family
of models that the fork supports. The purple bar corresponds to the SOTA XeTLA
int2 kernel~\cite{georganas2025} and the blue bar is BITCOS (this work), both
inside the vLLM XPU backend. The green number above each cluster of bars is
the speedup of BITCOS over the SOTA int2 kernel.}
\label{fig:e2egpu}
\end{figure*}

We integrated both the 2-bit LIBXSMM kernel~\cite{georganas2025} and the BITCOS GEMV kernels into the vLLM CPU
backend~\cite{vllm} and measured decode throughput on all three CPU platforms of
Section~\ref{sec:platforms} over 7 of the group-scaled LLM models of Table~\ref{tab:bits}.
We also benchmarked the same 7 models with the Prism ML fork of \texttt{llama.cpp}\footnote{\url{https://github.com/PrismML-Eng/llama.cpp}}. Two formats in that build are relevant here: \texttt{Q2\_0}, the fork's 2-bit code with
one fp16 scale per $128$ weights, which is the same bit rate and group size as
the SOTA 2-bit packing of prior work~\cite{georganas2025}, and
upstream's \texttt{TQ1\_0}, a five-trit per byte format.

Figure~\ref{fig:e2ecpu} shows the results of the end-to-end inference on the three CPU platforms. Each bar corresponds to the achieved decode throughput in tokens per second, for a single request of $256$ output tokens, with each inference engine at its own best thread count. The green number above each cluster of bars is the speedup of BITCOS over the SOTA 2-bit kernel, and we conclude that the two memory-bound platforms (EMR and ARL) benefit from BITCOS on every model, while the instruction-bound LNL CPU platform does not see any benefit, which is consistent with the roofline analysis in Section~\ref{sec:roofline}. We also make the following observations regarding the \texttt{llama.cpp} bars. First, the five-trit per byte format \texttt{TQ1\_0} is faster than the 2-bit \texttt{Q2\_0} on the two client platforms, by $1.66$--$1.68\times$ on Arrow Lake and $1.91$--$2.44\times$ on Lunar Lake, whereas on the server socket the two converge within $4\%$. It is worth noting that the SOTA 2-bit kernels of prior work~\cite{georganas2025} move more data than the \texttt{TQ1\_0} kernels of \texttt{llama.cpp}, and yet they outperform them by up to $1.78\times$. On the other hand, our BITCOS-based kernel outperforms the SOTA 2-bit kernel on the two memory-bound platforms in the range of $1.10$--$1.18\times$ on Emerald Rapids and $1.02$--$1.15\times$ on Arrow Lake. Compared to the \texttt{TQ1\_0} five-trit per byte format (which in principle is memory efficient), BITCOS is $1.13$--$1.48\times$ faster on Emerald Rapids and $1.46$--$1.74\times$ faster on Arrow Lake. On the instruction-bound LNL CPU platform the SOTA 2-bit work delivers the best end-to-end results and BITCOS loses on every model as predicted by the roofline model and the microbenchmarks of the previous section.

\subsection{Results on the Xe2 GPU platforms}
\label{sec:gpuresults}
We first present the GEMV microbenchmarks on each GPU platform (integrated GPU Arc 140V and discrete Arc Pro B70) and then the end-to-end decode results.

\subsubsection{GEMV microbenchmarks on the Arc 140V}
\label{sec:xe2}
To test the efficacy of the BITCOS-based Xe2 GEMV microkernel of Section~\ref{sec:decode-xe2} we use the same large 32768$\times$16384 ternary weight matrix as on the CPUs, group size 128, and a varying zero density $z$. Every BITCOS point is tuned independently over the candidate tiles and the int2 reference is tuned the same way. In Figure~\ref{fig:xe2shmoo}(a) top panel we illustrate the execution time of the GEMV on the Arc 140V, whereas on the bottom panel we show the corresponding effective bandwidth. We observe that the BITCOS format wins at every sampled density. In these plots we highlight with a green area the zero densities of interest, i.e.\ the $z\in[0.297,0.515]$ that the deployed checkpoints of Table~\ref{tab:bits} exhibit. For these zero densities, the observed speedup of the BITCOS-based GEMV over the int2 state-of-the-art GEMV is in the range of $1.04$--$1.14\times$, and the kernel delivers $71.7$--$75.6$\,GB/s of effective bandwidth across that band. The effective bandwidth does not stay flat but falls steadily with $z$, from $87.2$ to $65.5$\,GB/s across the full sweep, because the payload shrinks while the decode work per weight does not. This is why the realized speedup is smaller than the corresponding payload reduction.

\subsubsection{GEMV microbenchmarks on the Arc Pro B70}
\label{sec:b70}
We repeat the same GEMV benchmark on the discrete Arc Pro B70 (see Figure~\ref{fig:xe2shmoo}(b)) and the conclusions are largely the same as the ones on the Arc 140V: the BITCOS-based GEMV is never slower than the int2 reference. Over the same band of deployed zero densities, $z\in[0.297,0.515]$, the observed speedup is in the range of $1.01$--$1.12\times$, and the kernel holds $398.7$--$421.9$\,GB/s of effective bandwidth across that band. The effective bandwidth declines with $z$ for the same reason as on the integrated GPU platform, from $476.6$ to $342.6$\,GB/s. For example, at $z=0.40$ the measured speedup of $1.06\times$ (BITCOS vs 2-bit kernel) falls short of the $1.23\times$ byte ratio.

\subsubsection{End-to-end decode on the GPU platforms}
\label{sec:xe2-e2e}
We integrated both the int2 XeTLA kernel~\cite{georganas2025} and the BITCOS GEMV kernels into the same vLLM XPU backend~\cite{vllm} and measured decode throughput on both Xe2 platforms of
Section~\ref{sec:platforms} over the same 7 group-scaled LLM models of Table~\ref{tab:bits}. We also benchmarked the same 7 models with the Prism ML fork of \texttt{llama.cpp}, this time on its Vulkan backend, i.e.\ the vendor-neutral GPU path that is available in \texttt{llama.cpp}. Vulkan is the only backend of that fork which is available for the Xe2 GPUs, as the SYCL backend does not support the two relevant formats (\texttt{Q2\_0} and \texttt{TQ1\_0}). Of those two formats only \texttt{Q2\_0}, the 2-bit code with one fp16 scale per $128$ weights, has a Vulkan kernel. The five-trit per byte \texttt{TQ1\_0} format does not have any supporting Xe2 GPU kernel.

Figure~\ref{fig:e2egpu} shows the results of the end-to-end inference on the two Xe2 platforms. Each bar corresponds to the achieved decode throughput in tokens per second, for a single request of $256$ output tokens. The green number above each cluster of bars is the speedup of BITCOS over the SOTA int2 kernel, and we conclude that both Xe2 platforms benefit from BITCOS on every model, by $1.09$--$1.22\times$ on the integrated Arc 140V and $1.02$--$1.27\times$ on the discrete Arc Pro B70. We also make the following observations regarding the \texttt{llama.cpp} bars. At identical 2-bit format, identical group size and identical checkpoint the two 2-bit packing methods, (\texttt{Q2\_0} and the XeTLA int2 packing~\cite{georganas2025}) converge on the integrated GPU part within $5\%$ in the end-to-end inference results (Figure~\ref{fig:e2egpu}(a)). On the discrete GPU part (Figure~\ref{fig:e2egpu}(b)) the 2-bit XeTLA kernel delivers $1.46$--$2.12\times$ speedup over the \texttt{Q2\_0}-based inference. Our BITCOS-based inference outperforms the \texttt{Q2\_0} of \texttt{llama.cpp} by $1.11$--$1.18\times$ on the Arc 140V and by $1.61$--$2.30\times$ on the Arc Pro B70 and pushes the envelope of ternary LLM inference on Xe2 GPUs.

\section{Related Work}
\label{sec:related}

\subsection{Ternary Quantization of LLMs}
There are two main approaches to producing ultra-low-bit LLMs: quantization-aware training (QAT) and post-training quantization (PTQ). QAT applies the ternary-weights constraint during pretraining or fine-tuning. Early work established that binary and ternary weights are viable for natural language generation~\cite{binaryternarynlg}, and the seminal BitNet work~\cite{bitnet158,bitnet2b4t} showed that ternary $\{-1,0,+1\}$ weights keep the
accuracy of full-precision weights at scale. A subsequent work, ParetoQ~\cite{paretoq} showed that ternary and 2-bit formats live on the accuracy-size Pareto frontier, ahead of
1-bit and 4-bit formats, refining the earlier $k$-bit inference scaling laws that placed the optimum at 4 bits~\cite{kbitscaling,precisionscaling}. The Spectra/TriLM suite~\cite{trilm} pretrained
ternary models ranging from 99M to 3.9B parameters. A more recent QAT work, Tequila~\cite{tequila}, removes the trapping behavior that makes ternary QAT unstable by re-activating deadzone-trapped weights in the training process. Finally, more recent ternary QAT work has produced SOTA accuracy ternary LLMs for various LLM architectures: Bonsai~\cite{bonsai} models range from 1.7B to 27B parameters and offer multi-step reasoning, structured tool calls, vision tasks and agentic loops, while Maple~\cite{maple} is a 20B SOTA Mixture-of-Experts (MoE) ternary LLM with 1B active parameters. Such compact, high-quality models are a key enabler of on-device agentic systems~\cite{slmagentic}. Post-training quantization (PTQ) does not involve training steps and weight gradient updates. Instead, PTQ quantizes a pre-trained model and calibrates on a few sequences. Compared to QAT, PTQ is generally faster and less data-intensive, but may result in lower accuracy. Early PTQ methods at 4 and 2 bits, such as AWQ~\cite{awq} and QuIP~\cite{quip}, have not matched QAT at ultra-low bit-widths. Recent advancements in PTQ, CAT-Q~\cite{catq} and TWLA~\cite{twla}, make ternary weights directly from a pre-trained model and substantially narrow the accuracy gap compared to the QAT methods. These QAT and PTQ methods are complementary to our work: QAT and PTQ methods provide the means to obtain high-quality ternary LLM models, while our work yields memory-efficient and performant kernels to serve such ternary LLM inference on CPU and GPU platforms. 

\subsection{Kernels and runtimes for ternary LLM inference}

Bitnet.cpp~\cite{bitnetcpp} is the reference runtime for ternary LLMs and it is the companion runtime of the seminal ternary LLM work~\cite{bitnet158}. It is
faster than stock \texttt{llama.cpp}~\cite{llamacpp}, but recent work
showed that it does not deliver performance close to roofline~\cite{georganas2025}, thus in this work we compared against the SOTA runtime~\cite{georganas2025}. Earlier libraries targeted ternary and binary inference on edge devices by packing several low-bit weights per word and exploiting bit-serial or sign-based arithmetic, e.g.\ TernGEMM~\cite{terngemm} and TABv2~\cite{tabv2}. One related approach for ternary LLM kernels replaces multiplication with table lookup~\cite{maddness,lutnn,deepgemm}. T-MAC~\cite{tmac} pre-computes partial dot products, and uses the packed low-bit
codes as indices in a table lookup. This approach is applicable to CPUs with low vector FMA
throughput. On GPUs, prior work mixes 2-bit and 4-bit groups within a weight matrix and overlaps dequantization with the contraction to contain the accuracy loss~\cite{fast2bitgpu}. The 2-bit reference in this work is the LIBXSMM~\cite{libxsmm}/XeTLA~\cite{xetla} kernels~\cite{georganas2025}, and it is the strongest published baseline for CPU and Xe2 GPU
platforms to date. Section~\ref{sec:experiments} shows that the LIBXSMM/XeTLA kernels are equal to or faster
than the best \texttt{llama.cpp} configurations on all platforms. Therefore, the
reported gains of this work are measured against kernels that are roofline-optimal: the advantage of our work over these SOTA kernels stems from the fact that we exploit the inherent zero density of ternary LLM weights.

\subsection{Use of sparsity in ternary LLM inference}

Recent work~\cite{zhu2026sparsegemm} exploits sparsity in ternary LLMs by storing
only the indices of the non-zero weights in a Ternary CSC format, since the sign
alone describes a non-zero ternary weight, and replaces the multiplications with
additions and subtractions of the activations, which in theory reduces arithmetic by
$4\times$ at $50\%$ sparsity. This technique helps only in compute-bound cases:
prefill is compute-bound and becomes $1.2$--$2.3\times$ faster than libTorch on
Spectra TriLM 1.1B, but decode at batch size of one yields bandwidth-bound GEMV operations and the performance is even worse than the one of libTorch. The ternary CSC kernels also need
$75$--$88\%$ sparsity to overtake dense cuBLAS at the layer level, well above the $29.7$--$51.5\%$ that most
ternary checkpoints exhibit (Table~\ref{tab:bits}). Finally, the proposed index format has
a variable bit rate and reads the weights through irregular gathers, whereas our BITCOS work consists of a dense and positional bitmap, offering
sequential accesses and the resulting format is amenable to vectorizable unpacking.

Recent work makes the ternary sparsity \emph{semi-structured}, so that N:M
kernels can exploit it. Unlike the one-shot unstructured pruning of dense LLMs~\cite{sparsegpt}, the zeros here are produced by the quantizer itself. Sparse-BitNet~\cite{sparsebitnet} observes that the
$42\%$ zeros of a pretrained BitNet model are unstructured, and trains ternary
quantization jointly with a dynamic N:M mask, reporting up to $1.30\times$ end to
end speedup. Sherry~\cite{sherry} constrains every block of four weights to hold exactly
one zero, which leaves $4\times2^{3}=32$ distinct blocks that a five-bit code
stores exactly, i.e.\ $1.25$ bits per weight. Both these recent lines of work replace the distribution
that the quantizer produces and therefore require training: Sparse-BitNet must
apply the mask over the full pretraining run, while Sherry
fixes the zero density at $25\%$ and enforces a 3:4 pattern. Enforcing
semi-structured sparsity also costs accuracy: Sparse-BitNet reports that its 6:8
ternary models lose $0.17$--$0.32$ perplexity and $0.8$ to $3.8$ points of
downstream accuracy against their own dense ternary baselines. BITCOS instead uses the unstructured zeros that existing SOTA ternary LLM checkpoints already
contain. It is a change of the storage layout only, it applies to an existing
checkpoint, it needs neither retraining nor sparsity hardware, and it is
bit-exact, because it decodes the same ternary values as the existing/original ternary model. 

\section{Conclusion}

We introduced BITCOS, a distribution-adaptive ternary layout of a presence
bitmap plus a compacted sign vector that costs $2-z$ bits per weight at a zero
density $z$. Across 29 SOTA ternary LLM checkpoints the zero density ranges
from $29.7\%$ to $51.5\%$, so BITCOS stores 26 of them more compactly than the
five-trit packing and reaches $1.485$ bits per weight on the sparsest, while
against the 2-bit format in production it reduces weight traffic by
$1.16$--$1.32\times$ on all 29 models. Over the
zero densities these checkpoints exhibit, the BITCOS GEMV is $1.14$--$1.28\times$
faster than the state-of-the-art 2-bit kernel on a 64-core server Emerald Rapids CPU,
$1.13$--$1.27\times$ on a 24-core client Arrow Lake CPU, $1.04$--$1.14\times$ on the integrated Intel Xe2 Arc 140V GPU (Lunar Lake GPU) and
$1.01$--$1.12\times$ on the Arc Pro B70 discrete Intel GPU. End to end in vLLM over 7 ternary
LLMs, decode throughput improves by $1.10$--$1.18\times$, $1.02$--$1.15\times$,
$1.09$--$1.22\times$ and $1.02$--$1.27\times$ on the same four platforms. As future work we plan to extend the
BITCOS layout and its unpacking sequences to more CPU and GPU architectures.

\bibliographystyle{IEEEtran}
\bibliography{references}

\footnotesize
\noindent
\newline Optimization Notice: Software and workloads used in
performance tests may have been optimized for performance only on
Intel microprocessors.  Performance tests, such as SYSmark and
MobileMark, are measured using specific computer systems,
components, software, operations and functions.  Any change to any
of those factors may cause the results to vary.  You should
consult other information and performance tests to assist you in
fully evaluating your contemplated purchases, including the
performance of that product when combined with other products.
For more information go to \url{http://www.intel.com/performance}.

\noindent Intel, Xeon, and Intel Xeon Phi are trademarks of Intel Corporation in the U.S. and/or other countries.

\normalsize

\end{document}